\documentclass[runningheads, envcountsame, a4paper]{llncs}
\usepackage{graphicx}
\usepackage{comment}
\usepackage{amsmath,amssymb} 
\usepackage{color}
\usepackage{mathtools}  
\usepackage{bbm}  

\usepackage[misc]{ifsym}

\usepackage[hyphens]{url}

\usepackage{epsfig}

\usepackage{arydshln}

\usepackage{wrapfig}

\usepackage{subcaption}
\usepackage{algorithm} 
\usepackage{algorithmic}

\newcommand{\argmax}{\operatornamewithlimits{arg\,max}}
\newcommand{\argmin}{\operatornamewithlimits{arg\,min}}
\newcommand{\etal}{et al.}
\newcommand{\ie}{i.e.}
\newcommand{\eg}{e.g.}

\begin{document}
\title{Effective Version Space Reduction for Convolutional Neural Networks}
\titlerunning{Effective Version Space Reduction for ConvNets}
%
%
\author{Jiayu Liu \inst{1}\Letter \and 
        Ioannis Chiotellis \inst{1} \and
        Rudolph Triebel \inst{1,2} \and
        Daniel Cremers \inst{1}}
\authorrunning{J. Liu et al.}
%
%
\institute{
        Department of Informatics, Technical University of Munich, Munich, Germany
        \email{$\{$liuji,chiotell,triebel,cremers$\}$@in.tum.de} \and
        German Aerospace Center (DLR), Wessling, Germany}
%
\toctitle{Effective Version Space Reduction for Convolutional Neural Networks}
\tocauthor{Jiayu Liu, 
        Ioannis Chiotellis,
        Rudolph Triebel, and
        Daniel Cremers}

\maketitle              
\begin{abstract}
In active learning, sampling bias could pose a serious inconsistency problem and hinder the algorithm from finding the optimal hypothesis. However, many methods for neural networks are hypothesis space agnostic and do not address this problem. We examine active learning with convolutional neural networks through the principled lens of version space reduction. We identify the connection between two approaches---prior mass reduction and diameter reduction---and propose a new diameter-based querying method---the minimum Gibbs-vote disagreement. By estimating version space diameter and bias, we illustrate how version space of neural networks evolves and examine the realizability assumption. With experiments on MNIST, Fashion-MNIST, SVHN and STL-10 datasets, we demonstrate that diameter reduction methods reduce the version space more effectively and perform better than prior mass reduction and other baselines, and that the Gibbs vote disagreement is on par with the best query method.

\end{abstract}
%
%
%
\section{Introduction}
\label{sec:introduction}
Active learning is a supervised learning framework in which the learner is given access to a pool or stream of unlabeled samples and is allowed to selectively query labels from an oracle (\eg, a human annotator). In each query round, the learner queries the labels of some unlabeled samples and trains on the augmented labeled set to obtain new classifiers. The goal is to learn a good classifier or \textit{hypothesis} using as few labels as possible. This setting is relevant in many real-world problems, where labeled data are scarce or expensive to obtain, but unlabeled data are cheap and abundant. 

\begin{figure}[h]
    \begin{center}
        \includegraphics[width=1.0\linewidth]{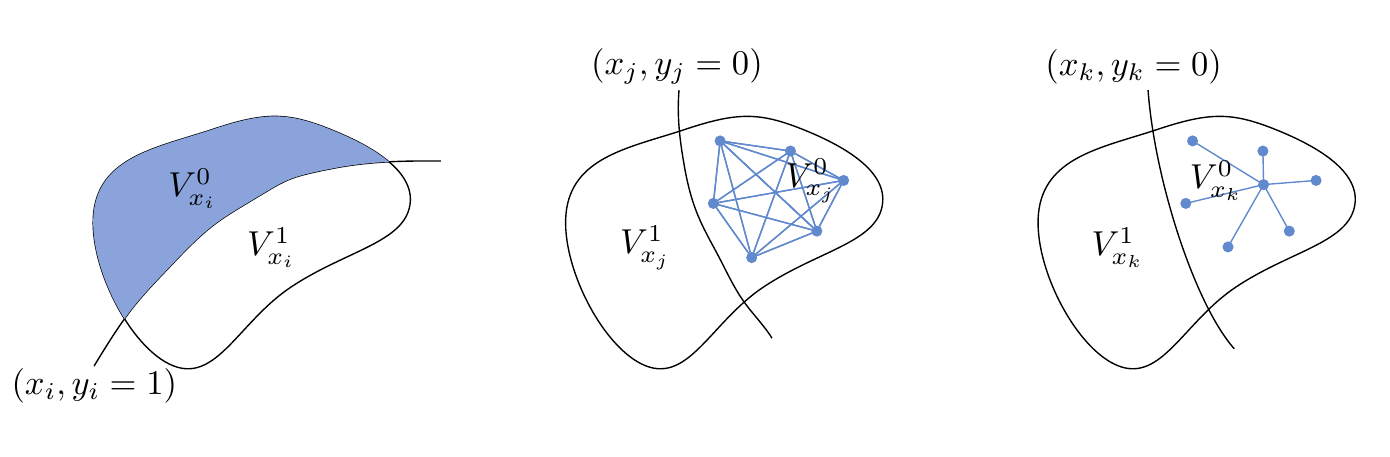}
    \end{center}
    \caption{Version space reduction for binary classification. Upon observing the label of $x$, the current version space $V\,$
    is split into subspaces $V_x^0$ and $V_x^1$, one of which will be removed and the other remains. \textbf{Left}: Prior mass reduction methods remove approximately half of the mass. \textbf{Middle}: Diameter reduction methods, like pairwise disagreement, query a sample that lead to sub-spaces of small diameter. \textbf{Right}: Proposed method, the \textit{Gibbs-vote disagreement}, measures diameter by the expected distance between random hypotheses and their majority vote.}
    \label{fig:pmr_vs_dr}
\end{figure}

Many active learning methods for neural networks rely on measures of the ``informativeness'' of a query, in the form of classifier uncertainty, margin \cite{Joshi09,Ducoffe18} or information gain \cite{Houlsby11,Gal17,Kirsch19}. Other methods capture the informativeness by representativeness of the query set using geometry-based \cite{Sener18} or discriminative \cite{Gissin19} methods. However, most of these methods ignore the notion of the hypothesis space and do not address the problem of sampling bias \cite{Dasgupta09}, which plague many active learning methods. Without carefully handling this problem, an active learning algorithm is not guaranteed to be \textit{consistent}, \ie, capable of finding the optimal classifier in the hypothesis space. 

We consider the hypothesis space of convolutional neural networks (ConvNets) and study version space reduction methods. 
Version space reduction works by removing hypotheses that are inconsistent with the observed labels from a predefined hypothesis space and maintaining the consistent sub-space, the \textit{version space}. A key condition called the \textit{realizability assumption} is that the hypothesis space contains the classifier that provides the ground truth---if not, there are no guarantees that the best hypothesis will not be removed, because a hypothesis might make mistakes on the queried samples but perform well on the data distribution. 

For neural networks, the realizability assumption may not hold for all cases. For instance, no neural networks can achieve arbitrarily small test error on some classification datasets. A workaround is to consider the effective labelings on a set of i.i.d. pool samples. To avoid the problem of an unreasonably large effective hypothesis space, as implied by the result of \cite{Zhang16}, we only consider the labelings achievable by training on unaltered samples and correct labels. 
We examine experimentally whether the realizability holds with this restriction and analyze its implications on version space reduction methods.

Prior mass reduction \cite{Dasgupta05,Golovin10,Cuong13} and diameter reduction \cite{Dasgupta06,Tosh17} are two widely used version space reduction approaches. See Fig.~\ref{fig:pmr_vs_dr} for illustration. However, prior mass reduction is not an appropriate objective for active learning 
\cite{Tosh17} since any intermediate version spaces containing more than one hypothesis may still have a large diameter, \ie, large error rate in the worst-case scenario, despite having substantially reduced mass. We derive connections between prior mass and diameter reduction and introduce a new interpretation of diameter reduction as prior mass ``reducibility reduction''.

We propose a new diameter measure called the \textit{Gibbs-vote disagreement}, which equals the expected distance between the random hypotheses and their majority vote classifier. We show its relation to a common diameter measure, the \textit{pairwise disagreement}, and discuss under which situations the former may be advantageous. We show experimentally on four image classification datasets that diameter reduction methods perform better than all baselines and that prior mass reduction \cite{Dasgupta05,Golovin10,Cuong13} and other baselines like \cite{Houlsby11,Gal17,Sener18,Ducoffe18} do not perform consistently better than random query and sometimes fail completely.

\section{Related Work}
\label{sec:related_work}

A lot of research has been conducted to study the label complexity for active learning and optimality guarantees for greedy version space reduction. Hanneke \cite{Hanneke07} and Balcan \etal~\cite{Balcan09} prove upper-bounds on the label complexity in the realizable and non-realizable cases, using a parameter called the disagreement coefficient. Tosh and Dasgupta \cite{Tosh17} propose a diameter-based active learning algorithm and characterize its sample complexity using a parameter called the splitting index. Dasgupta \cite{Dasgupta05} shows that a greedy strategy maximizing the worst-case prior mass reduction is approximately as good as the optimal strategy. Golovin and Krause \cite{Golovin10} show that the prior mass reduction utility function is adaptive submodular and a greedy algorithm is guaranteed to obtain near-optimal solutions in the average-case scenario. Cuong \etal~\cite{Cuong14} prove a worst-case optimality guarantee for pointwise submodular functions. 

A variety of methods relying on the informativeness of a query have been proposed for neural networks. Gal \etal~\cite{Gal17} use the Monte Carlo dropout to approximate the mutual information between predictions and model posterior \cite{Houlsby11} in a Bayesian setting. Kirsch \etal~\cite{Kirsch19} extend \cite{Houlsby11,Gal17} to a batch query method. Ducoffe and Precioso \cite{Ducoffe18} use adversarial attacks to generate samples close to the decision boundaries. Sener and Savarese \cite{Sener18} adopt a core-set approach to select representative samples for query. Gissin and Shalev-Shwartz \cite{Gissin19} use a discriminative method to select samples such that the labeled and the unlabeled set are indistinguishable. Pinsler \etal~\cite{Pinsler19} formulate batch query as a sparse approximation to the expected complete data posterior of model parameters in a Bayesian setting. Beluch \etal~\cite{Beluch18} show that ensemble methods consistently outperform geometry-based methods \cite{Sener18} and the Monte Carlo dropout method \cite{Gal16,Gal17}.

\section{Preliminaries}
\label{sec:preliminaries}

Let $\mathcal{X}$ be the input feature space and $\mathcal{Y}$ the label space. Let $\mathcal{H}$ be a hypothesis space of functions $h: \mathcal{X} \to \mathcal{Y}$ and assume a prior $\pi$ over $\mathcal{H}$. A hypothesis randomly drawn from the prior is called a \textit{Gibbs classifier}. Denote $S = \{(x_i, y_i)\}_{i=1}^n$ a pool of i.i.d. samples from the data distribution $P_{XY}$ and $Q \subseteq S$ the set of queried labeled samples. Define the version space $V$ corresponding to $Q$ as
\begin{align}
    V \coloneqq \left\{h\in \mathcal{H}: h(x) = y, \; \forall (x,y) \in Q \right\}.
\end{align}
Denote the subset of $V$ that is consistent with $x$ being labeled as $y$ as
\begin{align}
    V_x^y \coloneqq \left\{h\in \mathcal{H}: h(x) = y,\, h \in V \right\}.
\end{align}
The disagreement probability induced by the marginal distribution $P_X$ is defined as
\begin{equation}
d(h, h^\prime) \coloneqq \textmd{Pr}_{x} \! \left( h(x) \neq h^\prime(x) \right)
\end{equation}
which is a pseudo-metric on the hypothesis space. The disagreement and agreement region are defined as
\begin{gather}
    \text{DIS}(V) \coloneqq \left\{x \in \mathcal{X}: \exists h, h^\prime \in V, h(x) \neq h^\prime(x) \right\}, \\
    \text{AGR}(V) \coloneqq \mathcal{X} \setminus \text{DIS}(V).
\end{gather}

\section{Prior Mass Reduction}
\label{sec:prior_mass_reduction}

\subsection{Gibbs Error}
\label{sec:gibbs_error}

The Gibbs error \cite{Cuong13} of an unlabeled sample $x$ is the average-case relative prior mass reduction:
\begin{align}
    \text{GE}(x | V) \coloneqq\: \mathbb{E}_y\! \left[ 1 - \textmd{Pr}_{h \sim \pi|_V}(h(x)=y) \right] 
    =\: \mathbb{E}_y\! \left[ 1 - \pi|_V(V_x^y) \right],
    \label{eq:gibbs_error}
\end{align}
where $\pi|_V(h) = \pi(h)/\pi(V)$ is the conditional distribution of $\mathcal{H}$ restricted to $V$.
Gibbs error measures the proportion of inconsistent hypotheses taking expectation over all possible labelings of $x$, achievable by hypotheses in the version space.
A greedy strategy that considers maximizing the average-case absolute prior mass reduction in each query can equivalently select the unlabeled sample that maximizes the Gibbs error
\begin{align}
    \argmax_x \text{GE}(x | V).
    \label{eq:max_gibbs_error}
\end{align}

Define the prior mass reduction utility function as
\begin{align}
    f(Q) \coloneqq\: 1 - \textmd{Pr} \left( \left\{h\in \mathcal{H}: h(x) = y, \; \forall (x,y) \in Q \right\} \right) 
    =\: 1 - \pi(V).
    \label{eq:prior_mass_reduc_util_func}
\end{align}
The optimization problem in \eqref{eq:max_gibbs_error} can be written, up to a scaling factor, as
\begin{align}
    \argmax_x \pi(V) \text{GE}(x | V) 
    &= \argmax_x \mathbb{E}_y\! \left[ f(Q \cup \{(x,y)\}) - f(Q) \right] \\
    &= \argmax_x \Delta_{\text{avg}} (x | Q),
\end{align}
where the notation $\Delta_{\text{avg}}(x | Q)$ denotes the expected marginal gain of $x$ in terms of prior mass reduction given the labeled samples in $Q$.

A closely related objective for active learning is the label entropy given $x$. 
It can be shown that the Gibbs error lower bounds the entropy. However, a greedy strategy that maximizes the entropy is not guaranteed to 
be near-optimal in the adaptive case \cite{Cuong14}. Furthermore, empirically this criterion performs similarly or worse than the maximum Gibbs error. For the sake of simplicity, we do not consider this method in this paper.

\subsection{Variation Ratio}
\label{sec:variation_ratio}
The variation ratio of an unlabeled sample $x$ is the worst-case relative prior mass reduction upon the reveal of its label:
\begin{align}
    \text{VR}(x | V) \coloneqq& \: \min_y\left[ 1 - \textmd{Pr}_{h \sim \pi|_V}\! (h(x)=y) \right] 
    =\: \min_y\left[ 1 - \pi|_V(V_x^y) \right].
    \label{eq:var_ratio}
\end{align}
It measures the proportion of inconsistent hypotheses considering the worst-case labeling of $x$ and is a lower bound on the Gibbs error. A greedy strategy that considers maximizing the worst-case absolute prior mass reduction in each query selects the unlabeled sample that maximizes the variation ratio
\begin{align}
    \argmax_x \text{VR}(x | V),
    \label{eq:max_var_ratio}
\end{align}
which can be expressed in terms of the prior mass reduction utility function, up to a scaling factor, as
\begin{align}
    \argmax_x \pi(V) \text{VR}(x | V) 
    &= \argmax_x \min_y\left[ f(Q \cup \{(x,y)\}) - f(Q) \right] \\
    &= \argmax_x \Delta_{\text{wc}}\left(x | Q\right),
\end{align}
where the notation $\Delta_{\text{wc}}\left(x | Q\right)$ denotes the worst-case marginal gain of $x$ in terms of prior mass reduction given the labeled samples in $Q$.

\section{Diameter Reduction}
\label{sec:diameter_reduction}

\subsection{Worst-Case Pairwise Disagreement}
\label{sec:worst_case_pairwise_disagreement}
The size of the version space can be measured by the expected pairwise disagreement between hypotheses drawn from the conditional distribution:
\begin{align}
    \text{PWD}(V) \coloneqq\: \mathbb{E}_{h,h^\prime \sim \pi|_V}\! \left[ d(h, h^\prime) \right].
\end{align}
It is the \textit{average diameter} of the version space.
A greedy strategy selects the unlabeled sample that minimizes the worst-case pairwise disagreement
\begin{align}
    \argmin_x \max_y \text{PWD}(V_x^y) =\: \argmin_x \max_y \mathbb{E}_{h,h^\prime \sim \pi|_{V_x^y}}\! \left[ d(h, h^\prime) \right].
\end{align}
Other measures of diameter based on the supremum distance \cite{Kaariainen05,Dasgupta06} are not amenable to implementation because evaluation of such diameters involves optimization. The pairwise disagreement can be estimated from a finite set of sample hypotheses from the version space.

\subsection{Worst-Case Gibbs-Vote Disagreement}
\label{sec:worst_case_gibbs_vote_disagreement}
We propose to use a new diameter measure called the \textit{Gibbs-vote disagreement}. It is the expected disagreement between random hypotheses and their majority vote:
\begin{align}
    \text{GVD}(V) \coloneqq\: \mathbb{E}_{h \sim \pi|_V}\! \left[ d(h, h_{\text{vote}}|_V) \right],
\end{align}
where $h_{\text{vote}}|_V$ is the majority vote classifier of hypotheses from $V$. For each $x$, it induces a prediction
\begin{align}
    h_{\text{vote}}|_V(x) = \argmax_y \mathbb{E}_{h \sim \pi|_V}\! \left[ p(y|x; h) \right],
\end{align}
where $p(y|x;h)$ is the predicted probability of $x$ belonging to class $y$ given by a hypothesis $h$.
The majority vote classifier is the deterministic classifier that has the smallest expected distance to the Gibbs classifier \cite{Kaariainen05,Devroye13}:
\begin{align}
    \mathbb{E}_{h^\prime} \left[ d(h^\prime, h_{\text{vote}}) \right] = \min_h \mathbb{E}_{h^\prime}\! \left[ d(h^\prime, h) \right].
\end{align}
Hence the Gibbs-vote disagreement measures the size of the version space by the expected distance of the random hypotheses to their ``center''. Further, the following relation holds
\begin{align}
    \frac{1}{2}\, \text{PWD}(V) \le \text{GVD}(V) \le \text{PWD}(V)
    \label{eq:sandwich_pwd_gvd}
\end{align}
We defer the proof to the appendix. Essentially, Equation~\eqref{eq:sandwich_pwd_gvd} reveals that the Gibbs-vote disagreement is sandwiched between the average radius and diameter.

A greedy strategy selects the unlabeled sample that minimizes the worst-case Gibbs-vote disagreement
\begin{align}
    \argmin_x \max_y \text{GVD}(V_x^y) =\: \argmin_x \max_y \mathbb{E}_{h \sim \pi|_{V_x^y}}\! \left[ d(h, h_{\text{vote}}|_{V_x^y}) \right],
\end{align}
where $h_{\text{vote}}|_{V_x^y}$ is the majority vote of hypotheses from the subspace $V_x^y$ of the current version space if $x$ is labeled $y$.

\subsection{Diameter Reduction as Reducibility Reduction}
\label{sec:diameter_reduction_as_reducibility_reduction}
Pairwise disagreement shares a simple relation with Gibbs error---it is the expected Gibbs error:
\begin{align}
    \text{PWD}(V)
    &= \mathbb{E}_{h,h^\prime \sim \pi|_V}\! \left[\, \mathbb{E}_x\! \left[ \mathbbm{1}(h(x) \neq h^\prime(x)) \right]\, \right] \\
    &= \mathbb{E}_x\! \left[\, \mathbb{E}_{h \sim \pi|_V}\! \left[\, \textmd{Pr}_{h^\prime \sim \pi|_V}(h(x) \neq h^\prime(x)) \right]\, \right] \\
    &= \mathbb{E}_x\! \left[\, \text{GE}(x|V) \right].
    \label{eq:pw_ge}
\end{align}
A similar relation holds between Gibbs-vote disagreement and the variation ratio:
\begin{align}
    \text{GVD}(V) 
    &= \mathbb{E}_{h \sim \pi|_V}\! \left[\, \mathbb{E}_x\! \left[ \mathbbm{1}(h(x) \neq h_{\text{vote}}|_V\! (x)) \right]\, \right] \\
    &= \mathbb{E}_x\! \left[\,  \mathbb{E}_{h \sim \pi|_V}\! \left[ \mathbbm{1}(h(x) \neq h_{\text{vote}}|_V\! (x)) \right]\, \right] \\
    &= \mathbb{E}_x\! \left[\, \text{VR}(x|V) \right],
    \label{eq:gv_vr}
\end{align}
where the last equality holds because the predictions of the majority vote classifier are always the worst-case labels for prior mass reduction. Diameter reduction selects samples such that, upon revealing their labels, the induced subspaces have minimum possibility to be further reduced by a potential random query. Thus, it can be thought of as reducing the expected prior mass ``reducibility''. 

Prior mass reduction finds splits in directions that evenly partition the version space, but could result in version spaces that have irregular shapes, in the sense that the space can be whittled down finely in some directions while being under-split in others. The worst-case error rate of the resulted version space could still be large. Diameter reduction correctly resolve this issue. Fig.~\ref{fig:pmr_vs_dr} illustrates the differences between prior mass and diameter reduction.

\subsection{Weighted Diameter Reduction}
\label{sec:weighted_diameter_reduction}

Tosh and Dasgupta \cite{Tosh17} show that in general average diameter cannot be decreased at steady rate and propose to query the unlabeled samples that minimize the diameter weighted by the \textit{squared} prior mass in the worst-case scenario
\begin{align}
    \MoveEqLeft \argmin_x \max_y \mathbb{E}_{h,h^\prime \sim \pi}\! \left[ \mathbbm{1}(h, h^\prime \in V_x^y) \, d(h, h^\prime) \right] \\
    &= \argmin_x \max_y \pi(V_x^y)^2\, \mathbb{E}_{h,h^\prime \sim \pi|_{V_x^y}}\! \left[ d(h, h^\prime) \right].
\end{align}
The potential to be minimized is a surrogate for the ``amount'' of edges between hypotheses and is closely related to the splittablity of version space \cite{Tosh17,Dasgupta06}.

\section{Realizability Assumption}
\label{sec:realizability_assumption}

Even though neural networks are capable of fitting an arbitrary pool set, we show experimentally that the version space obtained by training on a subset of the pool set with stochastic gradient descent---the ``samplable'' version space---is biased and not likely to contain the correct labeling of the pool set. Indeed, the distance from the \textit{Bayes classifier}, which provides the ground truth labeling, to the ``boundary'' of the version space is non-negligible.

\begin{figure}[t!]
\begin{subfigure}{0.45\textwidth}
    \center
    \includegraphics[width=0.7\linewidth]{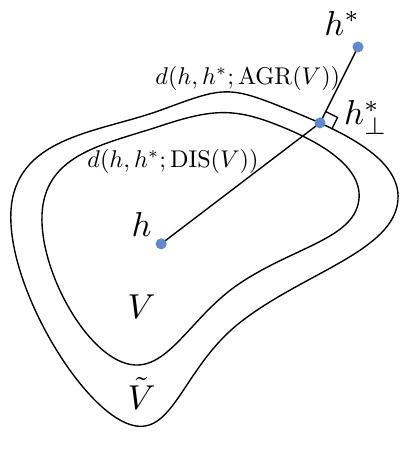}
\end{subfigure}%
\begin{subfigure}{0.55\textwidth}
    \centering
    \includegraphics[width=1.05\linewidth]{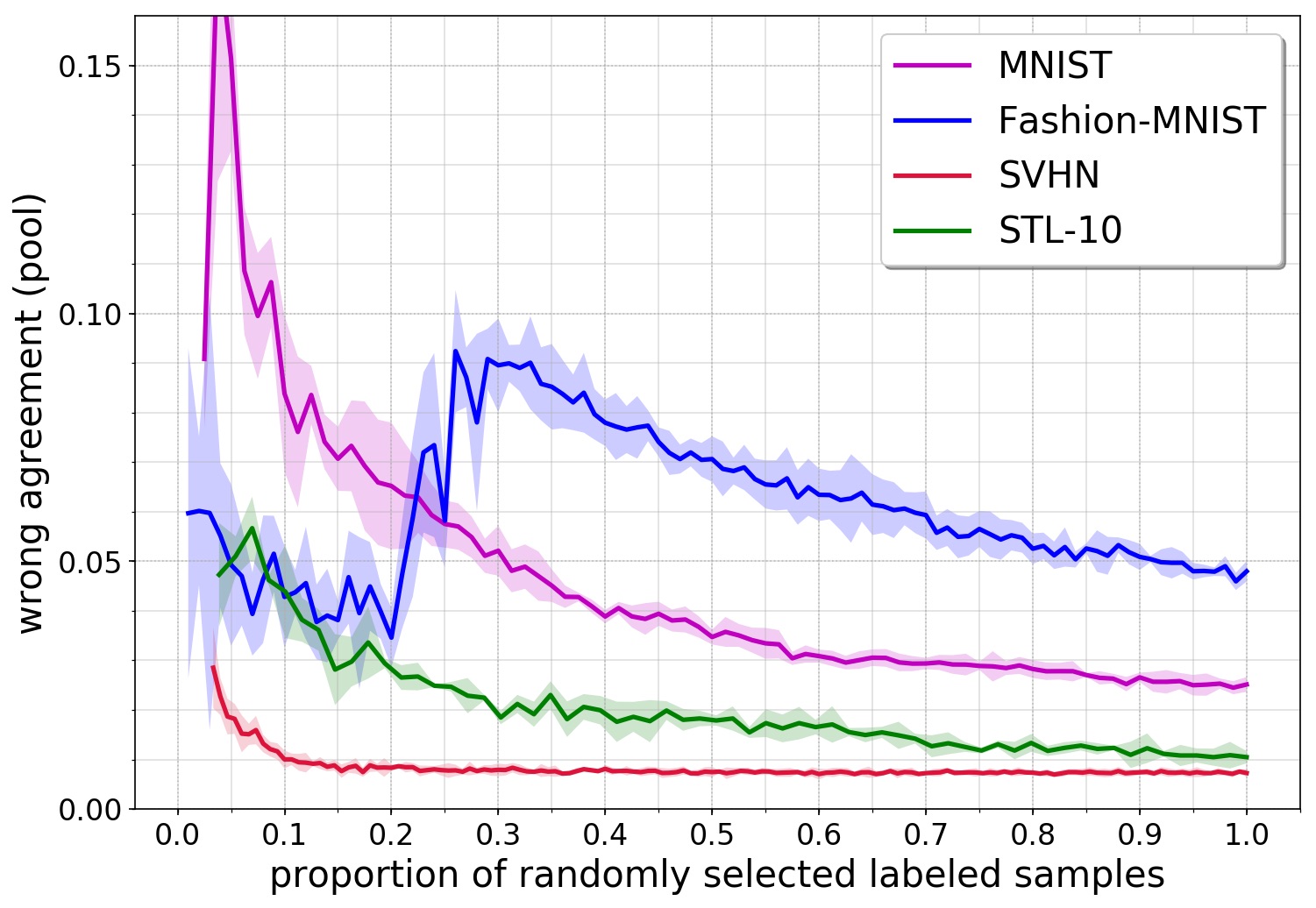}
\end{subfigure}
\caption{\textbf{Left}: Projection of $h^*$ to the samplable version space. \textbf{Right}: Wrong agreement of version spaces trained on random samples. Total numbers of samples are 400, 1000, 3000 and 2580 for MNIST, Fashion-MNIST, SVHN and STL-10 respectively.}
\label{fig:dist_decomp_wrg_agr}
\end{figure}

Let $h_\perp^*$ be the projection of the Bayes classifier $h^*$ to the set of hypotheses $\Tilde{V}$ that agree with $V$ on $\text{AGR}(V)$ (see the left plot of Fig.~\ref{fig:dist_decomp_wrg_agr}), \ie,
\begin{gather}
    h_\perp^* \coloneqq \argmin_{h \in \Tilde{V}} d(h, h^*), \\
    \Tilde{V} \coloneqq \left\{ h: h(\text{AGR(V)}) = h^\prime(\text{AGR}(V)), h^\prime \in V \right\}.
\end{gather}
It is easy to see that $h_\perp^*$  provides the ground truth on $\text{DIS}(V)$ and predicts the same labels on $\text{AGR}(V)$ as hypotheses in $V$ do, hence 
\begin{align}
    d(h_\perp^*, h^*) = d(h, h^*; \text{AGR(V)}) = \mathbb{E}_x\! \left[ \mathbbm{1}(x \in \text{AGV}(V))\mathbbm{1}(h(x) \neq h^*(x)) \right],\: \forall h \in V.
\end{align}
where $d(h, h^*; \text{AGR(V)})$ is the disagreement probability restricted to $\text{AGR(V)}$, or equivalently the \textit{wrong agreement} of hypotheses in $V$. 

\begin{figure}[t!]
\begin{center}
\includegraphics[width=0.7\linewidth]{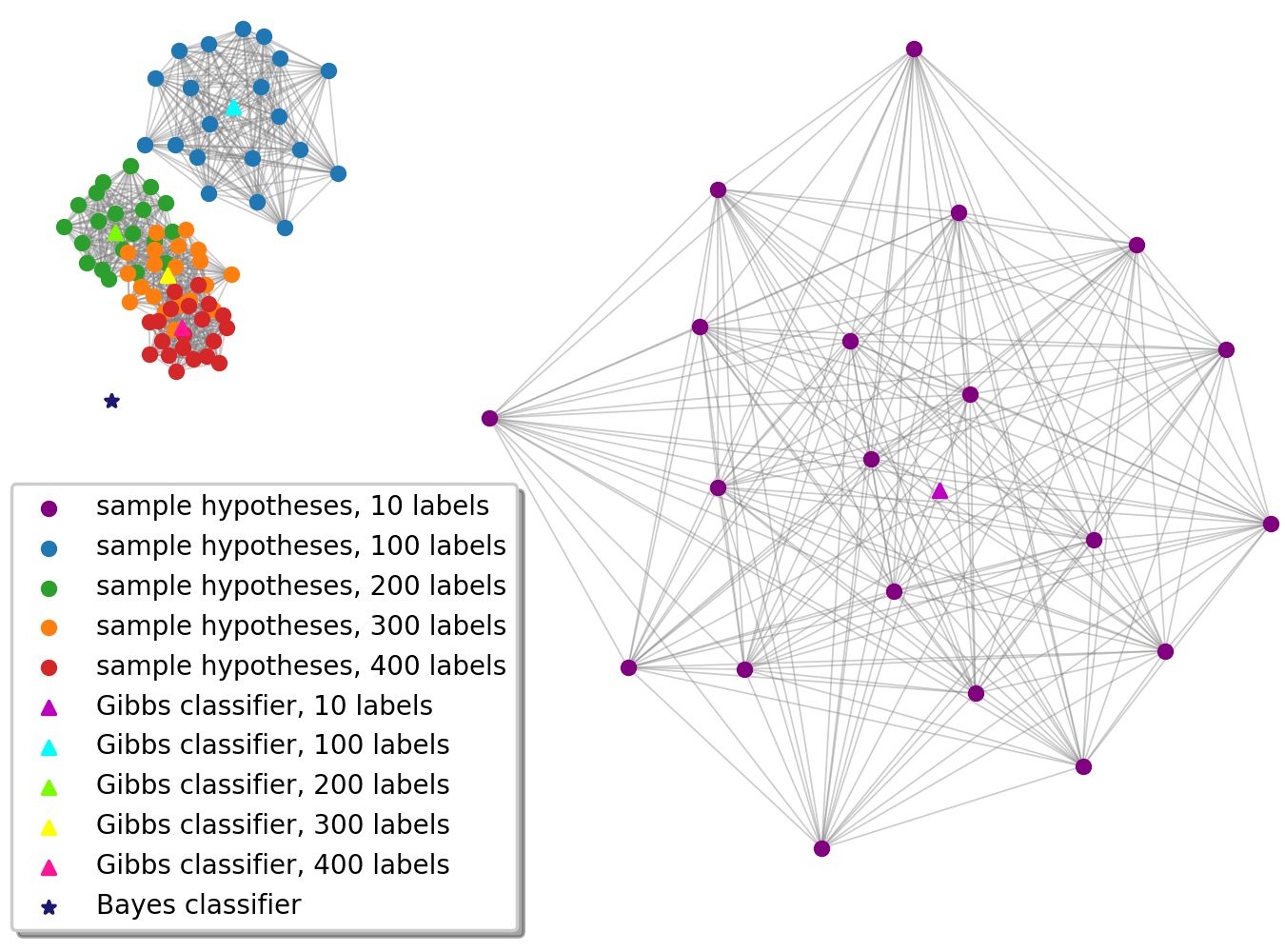}
\end{center}
   \caption{Embedding of version spaces on MNIST using MDS. As more random samples are used for training, the samplable version spaces move closer to the Bayes classifier but hardly cover it.}
\label{fig:mnist_mds_vs}
\end{figure}

We show the evolution of wrong agreement in the right plot of Fig.~\ref{fig:dist_decomp_wrg_agr}. As more random samples are queried, the wrong agreement decreases for all datasets, but for some much slower than the others. In Fig.~\ref{fig:mnist_mds_vs}, we show for MNIST a 2-D embedding of version spaces using Multi-Dimensional Scaling (MDS) \cite{Kruskal78}, which finds a low-dimensional representation of potentially high-dimensional data by preserving pairwise distances between the data points. The Bayes classifier is not contained in any of the samplable version spaces although the distances between them decrease steadily. 

In general neural networks trained with a random subset do not automatically predict all labels in the pool set correctly, unless a relatively large proportion of samples are used for training. However, this fact does not render version space reduction inconsistent, because the samplable version space is not fixed, but it shifts towards the correct labeling and finally covers it when the whole pool set has been used.

We conjecture that the dynamics of active learning with neural networks have two major components: (1) shrinkage of the samplable version space, which is explicitly optimized by the learning algorithm and (2) reduction of bias, which is not directly controllable. Empirical evidence is provided in the next section.

\section{Evaluation}
\label{sec:evaluation}
\noindent\textbf{Datasets and Architectures} 
We conduct active learning experiments\footnote{Source code is available at https://github.com/jiayu-liu/effective-version-space-reduction-for-convnets.} on four image classification datasets: MNIST, Fashion-MNIST, SVHN and STL-10.
Neural network architectures are chosen to be competent for each dataset but as simple as possible in the hope of controlling the model complexity and mitigating the effect of overfitting.
See Table~\ref{tab:set_al} for the complete experiment settings. \\

\noindent\textbf{Active Learning Methods} 
We compare nine querying methods: Random, variation ratio (VR), Gibbs error (GE), Bayesian Active Learning by Disagreement with Monte Carlo dropout (BALD-MCD) \cite{Houlsby11,Gal17}, Core-Set \cite{Sener18}, Deep-Fool Active Learning (DFAL) \cite{Ducoffe18}, pairwise disagreement (PWD), Gibbs-vote disagreement (GVD), 
and double-weighted pairwise disagreement (M$^2$-PWD) \cite{Tosh17}. For each method on each dataset, at least three runs of active learning with different random balanced initial training set are performed. \\

\begin{table*}[b]
\caption{Settings for each dataset used in the active learning experiments.}
\begin{center}
	\resizebox{\linewidth}{!}{
	\begin{tabular}{c|c|c|c|c|c}
	\hline
    Dataset & Pool/Val/Test & Model & Ensemble Size & Init/Query/Total & Runs \\
    \hline
    MNIST & 45000/5000/10000 & 2-conv-layer ConvNet & 20 & 10/5/400 & 4 \\
    Fashion-MNIST & 55000/5000/10000 & 3-conv-layer ConvNet & 20 & 10/10/1000 & 4\\
    SVHN & 40000/5000/15000 & 6-conv-layer ConvNet & 20 & 100/20/3000 & 4 \\
    STL-10 & 4000/1000/8000 & ResNet18 & 20  & 100/40/2580 & 3  \\
    \hline
	\end{tabular}}
\end{center}
\label{tab:set_al}
\end{table*}

\begin{figure*}[t!]
	\begin{center}
		\includegraphics[width=1.0\linewidth]{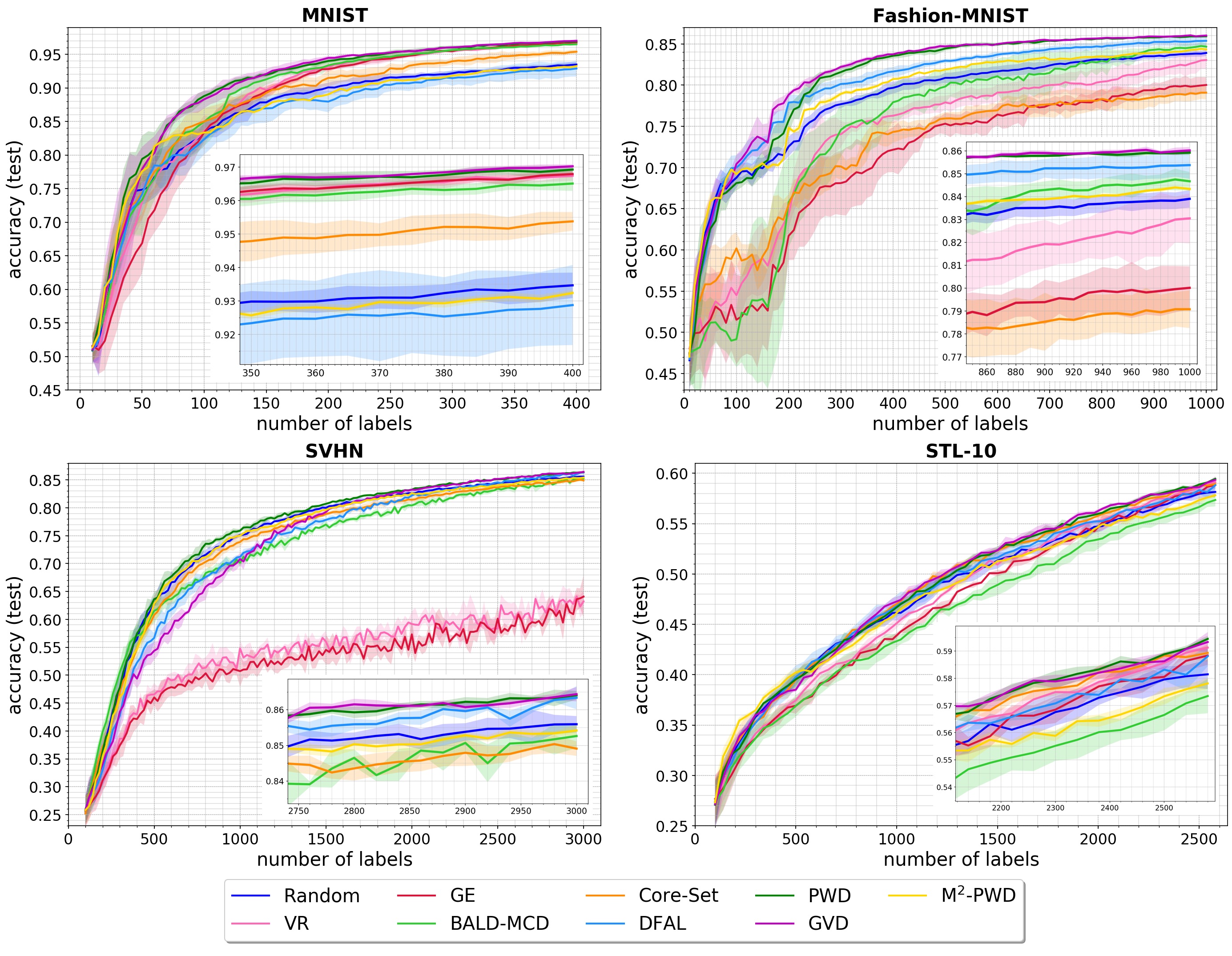}
	\end{center}
	\caption{Accuracy over number of queried labels on the test set. Direct diameter reduction methods PWD and GVD are consistently better than Random and are among the best methods. Weighted diameter reduction M$^2$-PWD is on par with Random. Other baselines are effective on some datasets but inferior to Random on the others. Note that PWD, GVD and M$^2$-PWD exhibit smaller variances than the others.}
	\label{fig:acc_te}
\end{figure*}

\noindent\textbf{Ensemble Size} 
We train networks multiple times from scratch to obtain sample hypotheses and use them for prior mass and diameter estimation. Since diameters are estimated by considering partitioned version spaces, the ensemble size should be at least in the order of number of classes. We set the size to 20. Larger ensemble improves estimation but at the cost of longer training time. In preliminary experiments, we tried larger ensembles (40) and did not observe significant differences. Hence we do not include experiments on changing this hyper-parameter in the paper. \\

\noindent\textbf{Query Size} 
We set a small query budget for each round to reduce the correlation between queries. Larger budget may alleviate the pressure of frequent retraining, but the effect of each query can not be estimated and examined reliably. We observed in preliminary experiments that using larger budget (one or two orders larger) hides the differences between methods.

\begin{figure*}[ht!]
\begin{center}
\includegraphics[width=1.0\linewidth]{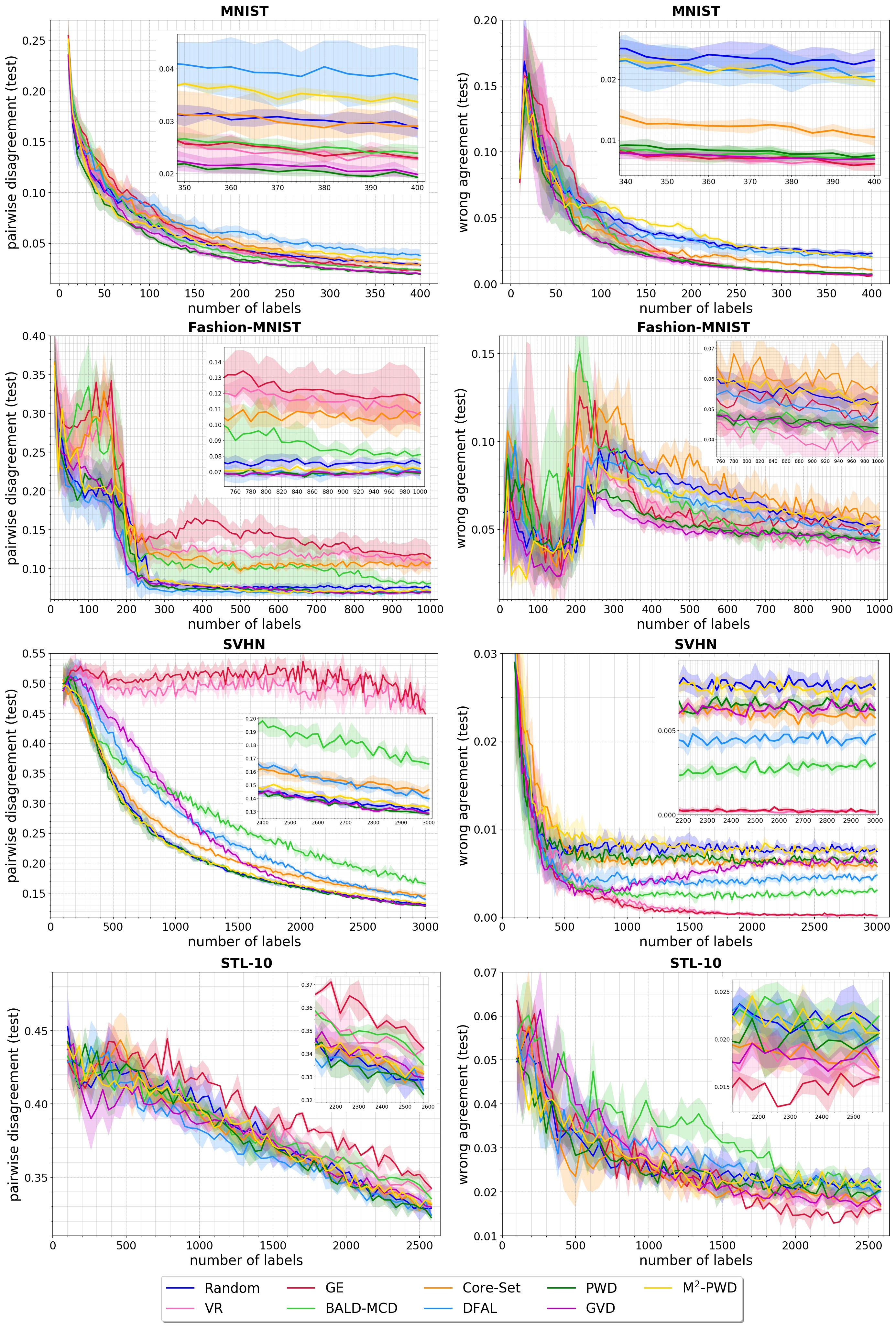}
\end{center}
   \caption{Pairwise disagreement and wrong agreement over number of queried labels on the test set. Except direct diameter reduction methods PWD and GVD, other baselines are not consistently better than or on par with Random at reducing version space diameter. Performing worse than Random: GE, VR and BALD-MCD on datasets except MNIST, Core-Set on Fashion-MNIST and SVHN, and DFAL on MNIST and SVHN, and M$^2$-PWD on MNIST.}
\label{fig:pw_dis_wrg_agr_te}
\end{figure*}



\begin{table}[b!]
\caption{Accuracy on the test set in percentage.}
\begin{center}
	\resizebox{0.85\linewidth}{!}{%
		\begin{tabular}{l|c|c|c|c}
		\hline
		& \textbf{MNIST} & \textbf{Fashion-MNIST} & \textbf{SVHN} & \textbf{STL-10} \\
		\#labels  & 400                       & 1000                      & 3000           & 2580 \\
		\hline
		Random    & 93.47 $\pm$ 0.38          & 83.90 $\pm$ 0.38          & 85.60 $\pm$ 0.23          & 58.15 $\pm$ 0.54 \\ 
		\hdashline
		VR        & 96.74 $\pm$ 0.15          & 83.05 $\pm$ 1.09          & 63.23 $\pm$ 1.99          & 59.13 $\pm$ 0.21 \\
		GE        & 96.79 $\pm$ 0.10          & 80.01 $\pm$ 0.94          & 64.08 $\pm$ 3.77          & 58.84 $\pm$ 0.34 \\
		\hdashline
		BALD-MCD  & 96.51 $\pm$ 0.22          & 84.67 $\pm$ 0.41          & 85.26 $\pm$ 0.34          & 57.35 $\pm$ 0.64 \\
		Core-Set  & 95.38 $\pm$ 0.28          & 79.08 $\pm$ 0.82          & 84.91 $\pm$ 0.20          & 58.93 $\pm$ 0.33 \\
		DFAL      & 92.88 $\pm$ 1.19          & 85.38 $\pm$ 0.60          & 86.34 $\pm$ 0.33          & 58.81 $\pm$ 0.37 \\
		\hdashline
		PWD       & 96.92 $\pm$ 0.12          & 85.92 $\pm$ 0.10          & 86.41 $\pm$ 0.12          & \textbf{59.45} $\pm$ \textbf{0.11} \\
		GVD       & \textbf{97.02} $\pm$ \textbf{0.06} & \textbf{86.01} $\pm$ \textbf{0.15} & \textbf{86.44} $\pm$ \textbf{0.20} & 59.33 $\pm$ 0.37 \\
		M$^2$-PWD & 93.24 $\pm$ 0.09         & 84.33 $\pm$ 0.03           & 85.42 $\pm$ 0.16          & 57.81 $\pm$ 0.20 \\
		\hline
		\end{tabular}%
	}
\end{center}
\label{tab:test_acc}
\end{table}



\begin{table}[b!]
\caption{Diameter (pairwise disagreement)  on the test set in percentage.}
\begin{center}
	\resizebox{0.85\linewidth}{!}{%
		\begin{tabular}{l|c|c|c|c }
		\hline
		& \textbf{MNIST} & \textbf{Fashion-MNIST} & \textbf{SVHN} & \textbf{STL-10} \\
		\#labels  & 400           & 1000          & 3000            & 2580 \\
		\hline
		Random    & 2.86 $\pm$ 0.18          & 7.55 $\pm$ 0.26   		& 13.13 $\pm$ 0.29           & 32.88 $\pm$ 0.43 \\
		\hdashline
		VR        & 2.27 $\pm$ 0.18          & 10.64 $\pm$ 0.72         & 46.88 $\pm$ 2.76           & 34.21 $\pm$ 0.08 \\
		GE        & 2.30 $\pm$ 0.04          & 11.38 $\pm$ 1.52    	    & 44.87 $\pm$ 4.25   		 & 34.25 $\pm$ 0.09 \\
		\hdashline
		BALD-MCD  & 2.39 $\pm$ 0.15          & 8.11 $\pm$ 0.51          & 16.58 $\pm$ 0.42           & 33.55 $\pm$ 0.44 \\
		Core-Set  & 2.91 $\pm$ 0.18          & 10.79 $\pm$ 1.34         & 14.66 $\pm$ 0.47           & 33.13 $\pm$ 0.64 \\
		DFAL      & 3.79 $\pm$ 0.60          & 7.06 $\pm$ 0.60          & 13.98 $\pm$ 0.31           & 32.41 $\pm$ 0.27 \\
		\hdashline
		PWD       & \textbf{1.93} $\pm$ 0.04 & \textbf{6.91} $\pm$ 0.16 & \textbf{12.80} $\pm$ 0.08  & \textbf{32.25} $\pm$ 0.26 \\
		GVD       & 1.98 $\pm$ 0.05		     & 6.98 $\pm$ 0.26		    & 12.88 $\pm$ 0.25           & 32.96 $\pm$ 0.48 \\
		M$^2$-PWD & 3.37 $\pm$ 0.13   	  	 & 7.22 $\pm$ 0.08          & 13.31 $\pm$ 0.13           & 33.23 $\pm$ 0.18 \\
		\hline
		\end{tabular}%
		}
\end{center}
\label{tab:diam_reduc_te}
\end{table}

\subsection{Diameter Reduction is More Effective Than Prior Mass Reduction}
\label{sec:diameter_reduction_is_more_effective_than_prior_mass_reduction}
Fig.~\ref{fig:acc_te} and Table~\ref{tab:test_acc} show that direct diameter reduction methods PWD and GVD are consistently better than Random and achieve higher accuracy than other baselines while weighted diameter reduction M2-PWD is on par with Random. Diameter reduction methods usually exhibit less variances because training on samples queried by PWD, GVD and M2-PWD yields version spaces with smaller diameters and less diverse sample hypotheses. Prior mass reduction is not always effective and even fails on SVHN. This failure is an example of prior mass reduction being incapable of reducing the diameter, and provides empirical evidence that it may not be an appropriate objective for active learning. 

\subsection{Comparison to Other Baselines}
\label{sec:comparison_to_other_baselines}
BALD-MCD, Core-Set and DFAL are not consistently better than Random although each of them achieves comparative test accuracy on certain dataset. Their inferiority to Random in terms of test accuracy usually correlates with higher diameter (See description in Fig.~\ref{fig:pw_dis_wrg_agr_te} and Table~\ref{tab:diam_reduc_te}). BALD-MCD and DFAL are highly related to prior mass reduction methods in that BALD \cite{Houlsby11} seeks samples for which the model parameters under the posterior disagree the most about the prediction \cite{Houlsby11}, and that DFAL, inspired by margin-based active learning \cite{Balcan07}, tries to locate the decision boundary with fewer labels which is essentially removing inconsistent hypotheses in the realizable case. However, none of them explicitly minimize the diameter, neither does Core-Set. 

Note that for a fair comparison, we do not augment the training set by also adding the adversarial samples as the original DFAL paper \cite{Ducoffe18} does. Samples with minimum adversarial perturbation are then verified reliably to be less effective than those lead to minimum diameter. The original Core-Set paper \cite{Sener18} uses a large query batch size (in the order of 1000). However, many baselines rely on greedy selection and do not perform any batch optimization. To reduce query correlation, we adopt as small batch size as possible. This allows reliable evaluation of the effectiveness of queried samples as in the online setting. We are therefore able to identify one major cause of inferiority to Random as failing to effectively reduce the version space diameter. 

\subsection{Evolution of Samplable Version Space and its Implications}
\label{sec:evolution_of_samplable_version_space_and_its_implications}
As shown in Fig.~\ref{fig:pw_dis_wrg_agr_te} and \ref{fig:mnist_mds_vs},
the samplable version space shifts closer to the correct labeling while reducing its diameter as more labels are queried. These two processes together result in smaller test error. \\

\noindent\textbf{No Direct Control Over Reduction of Version Space Bias}
\label{para:active_learning_algorithms_have_no_direct_control_over_reduction of_version_space_bias}
Interestingly, the Core-Set method, which queries representative samples from the pool set by solving a k-center problem in the feature space learned by neural networks, is incapable of achieving negligible wrong agreement on the learned version spaces. Indeed, it suffers larger version space bias than the direct diameter reduction methods. After all, random queries which are i.i.d. by assumption fail to achieve this goal as concluded in Section~\ref{sec:realizability_assumption} and other attempts without augmenting the training data seem doomed. \\

\noindent\textbf{Prior Mass Induced by Stochastic Gradient Descent May Not Be a Reliable Surrogate Measure} 
\label{para:prior_mass_induced_by_stochastic_gradient_descent_may_not_be_a_reliable_surrogate_measure} 
The continued decline in wrong agreement indicates that the distribution over labelings changes over time. This fact of shifting density over samplable labelings renders the notion of prior mass problematic, hence all notions relying on prior mass may not be well-defined. A direct consequence is that an estimate of the worst-case version space reduction would be more reliable than the average-case one. For example, VR provides a more reliable estimate of version space reduction than GE does. \\ 

\noindent\textbf{Inferiority of Weighted Diameter Reduction Method}
\label{para:inferiority_of_weighted_diameter_reduction_methods}
The estimation of weighted diameter involves estimating the prior mass. Hence, the inferiority of M$^2$-PWD to PWD and GVD can be attributed to the intrinsic difficulty of obtaining unbiased samplable version spaces and the resulted density shift. A supportive evidence can be seen by noting that on MNIST and Fashion-MNIST, where the wrong agreement is large (hence large density shift), the weighted variant performs worse, while on SVHN and STL-10, where the wrong agreement is small (hence small shift), the gap is less significant.

\begin{wrapfigure}{R}{0.55\textwidth}
\includegraphics[width=0.55\textwidth]{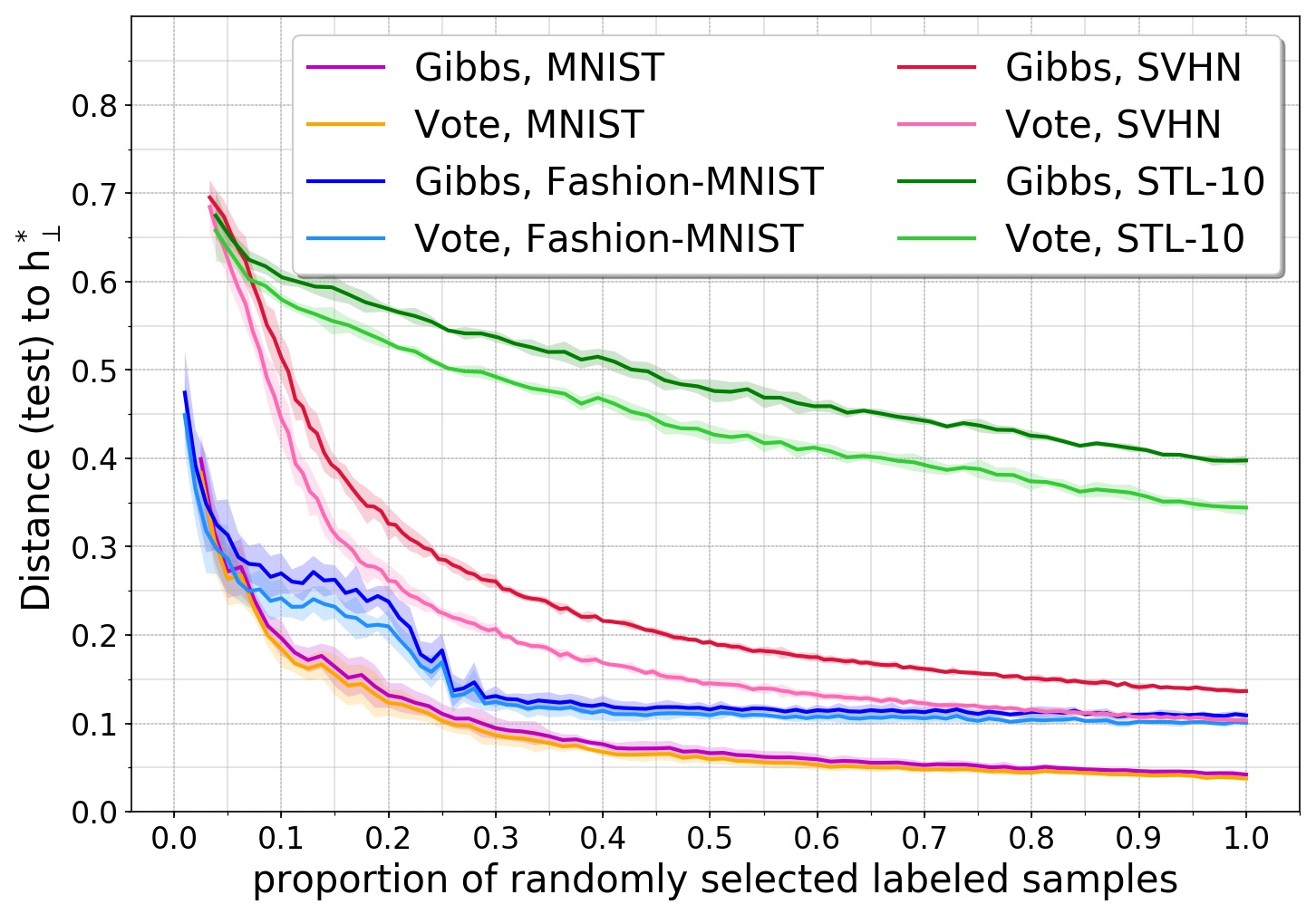}
\caption{Distance from the Gibbs and the majority vote classifier to the projection of $h^*$. On four datasets, the majority vote classifier has a smaller distance, hence smaller error rate. See description of Fig.~\ref{fig:dist_decomp_wrg_agr} for total numbers of random samples.}
\label{fig:dist_to_proj_rand_te}
\end{wrapfigure}

\subsection{Gibbs-Vote Disagreement}
\label{sec:gibbs_vote_disagreement}
The Gibbs-vote disagreement is among the best methods on all datasets, except for the early learning stage on SVHN. Its effectiveness can be ascribed to an interesting phenomenon---majority voting reduces mistakes. Although it need not necessarily be the case, this phenomenon occurs in many situations and the boost to accuracy depends on the variance of errors of Gibbs classifiers \cite{Lacasse07}. We show empirically that the majority vote classifier indeed has smaller error rate than random hypotheses in the version space in Fig.~\ref{fig:dist_to_proj_rand_te}. 
Hence, optimizing the Gibbs-vote disagreement not only reduces the diameter but also implicitly moves the consistent hypotheses closer to the correct labeling, which is useful when the samplable version spaces are biased and do not contain the Bayes classifier.

\section{Conclusion}
\label{sec:conclusion}
In this work, we studied version space reduction for convolutional neural networks. We revealed the differences and connections between prior mass and diameter reduction methods and proposed the Gibbs-vote disagreement as a new effective diameter-reduction method. With experiments on four datasets, we shed light into how version space reduction works in the deep active learning setting and demonstrated the superiority of diameter reduction over prior mass reduction methods and other baselines.

\section*{Acknowledgments}
This work was supported by the BMBF project MLWin and the Munich Center for Machine Learning (MCML).

\clearpage
%
%
%

\appendix

\section{Estimators and Algorithm}

\subsection{Effective Hypothesis Space}
Let $X_S$ be the set of unlabeled samples in the pool set $S$ and $X_Q$ the set of queried unlabeled samples. The effective hypothesis space $\hat{H}$ is the restriction of $\mathcal{H}$ to $X_S$, or equivalently all possible labelings of $X_S$: 
\begin{multline}
    \hat{\mathcal{H}} \coloneqq \mathcal{H}|_{X_S} = \{ ((h(x_1), h(x_2),\, \cdots,\, h(x_n)): \\
    \forall h \in \mathcal{H}, x_i \in X_S, 1\le i \le n) \}
\end{multline}

\subsection{Estimators of Diameters}
Let $\hat{V}$ be a set of $M$ sample hypotheses from $\pi|_V$
\begin{align}
     \hat{V} \coloneqq \{ h_m: h_m \stackrel{i.i.d.}{\sim} \pi|_V,\, 1 \le m \le M\}
\end{align}
Assuming $M = |\hat{V}| \ge 2$, an unbiased estimator of the pairwise disagreement can be constructed by computing average pairwise distances between hypotheses in $\hat{V}$
\begin{align}
    \frac{1}{|\hat{V}|(1 - |\hat{V}|)} \sum_{h \in \hat{V}}\sum_{h^\prime \in \hat{V},\, h^\prime \neq h} d(h, h^\prime)
\end{align}
Similarly, an unbiased estimator of the Gibbs vote disagreement can be constructed by computing average distances between hypotheses in $\hat{V}$ and the empirical majority vote $h_{\text{vote}}|_{\hat{V}}$
\begin{align}
    \frac{1}{|\hat{V}|} \sum_{h \in \hat{V}} d(h, h_{\text{vote}}|_{\hat{V}})
\end{align}

\subsection{Algorithm}
A (batch mode) greedy algorithm that selects the unlabeled samples which induce the minimum version space in terms of a given diameter measure in the worst-case scenario is shown in Algorithm \ref{alg:min_gv_dis}. Other active learning methods (\eg, prior mass reduction methods) can also be described using this algorithm with line $11$ in the algorithm replaced by the corresponding objective functions. Note that in line $2$ and $15$ the version space is maintained explicitly while in practice we only need to sample from the new version space by training neural networks on the updated set of queried samples. Hence, the version space is always implicitly maintained.

\begin{algorithm}
\caption{Worst-Case Diameter Reduction}
\label{alg:min_gv_dis}
\begin{algorithmic}[1]
\STATE \textbf{Input}: $T$ query rounds, $k$ batch size of query, $M$ size of ensemble, $X_S$ pool of unlabeled samples, 
$\text{diam}$ diamter measure, $\pi$ prior over hypotheses
\STATE $V_0 \leftarrow \hat{\mathcal{H}}$
\STATE $Q \leftarrow \varnothing$
\FOR{$t \leftarrow 1$ to $T$}
    \STATE $\hat{V} \leftarrow \varnothing$
    \FOR{$m \leftarrow 1$ to $M$}
        \STATE sample $h_m \sim \pi|_{V_t}$
        \STATE $\hat{V} \leftarrow \hat{V} \cup \{h_m\}$
    \ENDFOR
    \FOR{$j \leftarrow 1$ to $k$}
        \STATE $x_j \leftarrow \argmin\limits_{x \in X_S\setminus X_Q} \max\limits_{y \in \mathcal{Y}}  
        \,\text{diam}\left( \hat{V}_x^y \right)$
        \STATE query $y_j$
        \STATE $Q \leftarrow Q \cup \{ (x_j, y_j) \}$
    \ENDFOR
    \STATE $V_t \leftarrow \{ h\in V_{t-1}: h(x)=y,\, \forall\, (x, y) \in Q \}$ 
\ENDFOR
\RETURN any $h \sim \pi|_{V_t}$ 
\end{algorithmic}
\end{algorithm}


\section{Proof of Equation \ref{eq:sandwich_pwd_gvd}}
\begin{proof}

\begin{align*}
    \text{GVD}(V) &= \mathbb{E}_{h^\prime \sim \pi|_V}\! \left[ d(h^\prime, h_{\text{vote}|_V}) \right] \\
    &= \frac{1}{2} \min_h \left\{ \mathbb{E}_{h^\prime, h^{\prime\prime} \sim \pi|_V}\! \left[ d(h^\prime, h) + d(h^{\prime\prime}, h) \right] \right\} \\
    &\ge \frac{1}{2} \mathbb{E}_{h,h^\prime \sim \pi|_V}\! \left[ d(h, h^\prime) \right] \\
    &= \frac{1}{2}\, \text{PWD}(V).
\end{align*}
where the last inequality is by triangular inequality. By definition, it holds that $\text{VR}(x|V) \le \text{GE}(x|V)$. Using the relations derived in Section~\ref{sec:diameter_reduction_as_reducibility_reduction}
\begin{align*}
    \text{PWD}(V) &= \mathbb{E}_x\! \left[\, \text{GE}(x|V) \right], \\
    \text{GVD}(V) &= \mathbb{E}_x\! \left[\, \text{VR}(x|V) \right],
\end{align*}
it holds that
\begin{align*}
    \text{GVD}(V) \le \text{PWD}(V).
\end{align*}
\end{proof}

\section{Singly-Weighted Diameter Reduction}
Besides the doubly-weighted diameter reduction method mentioned in Section~\ref{sec:weighted_diameter_reduction}, there are singly-weighted variants. 
\subsubsection{Weighted Pairwise Disagreement} 
\begin{align}
    \argmin_x \max_y \pi(V_x^y)\, \mathbb{E}_{h,h^\prime \sim \pi|_{V_x^y}}\! \left[ d(h, h^\prime) \right],
\end{align}
which is equivalent to minimizing the potential expected marginal gain on the prior mass reduction utility function
\begin{align}
    \argmin_x \max_y \mathbb{E}_x\! \left[ \pi(V_x^y)\, \text{GE}(x | V_x^y) \right] 
    = \argmin_x \max_y \mathbb{E}_x\! \left[ \Delta_{\text{avg}}(x|Q \cup \{(x,y)\}) \right].
\end{align}

\subsubsection{Weighted Gibbs-Vote Disagreement} 
\begin{align}
    \argmin_x \max_y \pi(V_x^y)\, \mathbb{E}_{h \sim \pi|_{V_x^y}}\! \left[ d(h, h_{\text{vote}}|_{V_x^y}) \right],
\end{align}
which is equivalent to minimizing the potential worst-case marginal gain 
\begin{align}
    \argmin_x \max_y \mathbb{E}_x\! \left[ \pi(V_x^y)\, \text{VR}(x | V_x^y) \right] 
    = \argmin_x \max_y \mathbb{E}_x\! \left[ \Delta_{\text{wc}}(x|Q \cup \{(x,y)\}) \right].
\end{align}

\section{Additional Evaluation Results}
\subsection{Evaluation on the Pool Set}
\begin{figure*}[ht!]
\begin{center}
\includegraphics[width=1.0\linewidth]{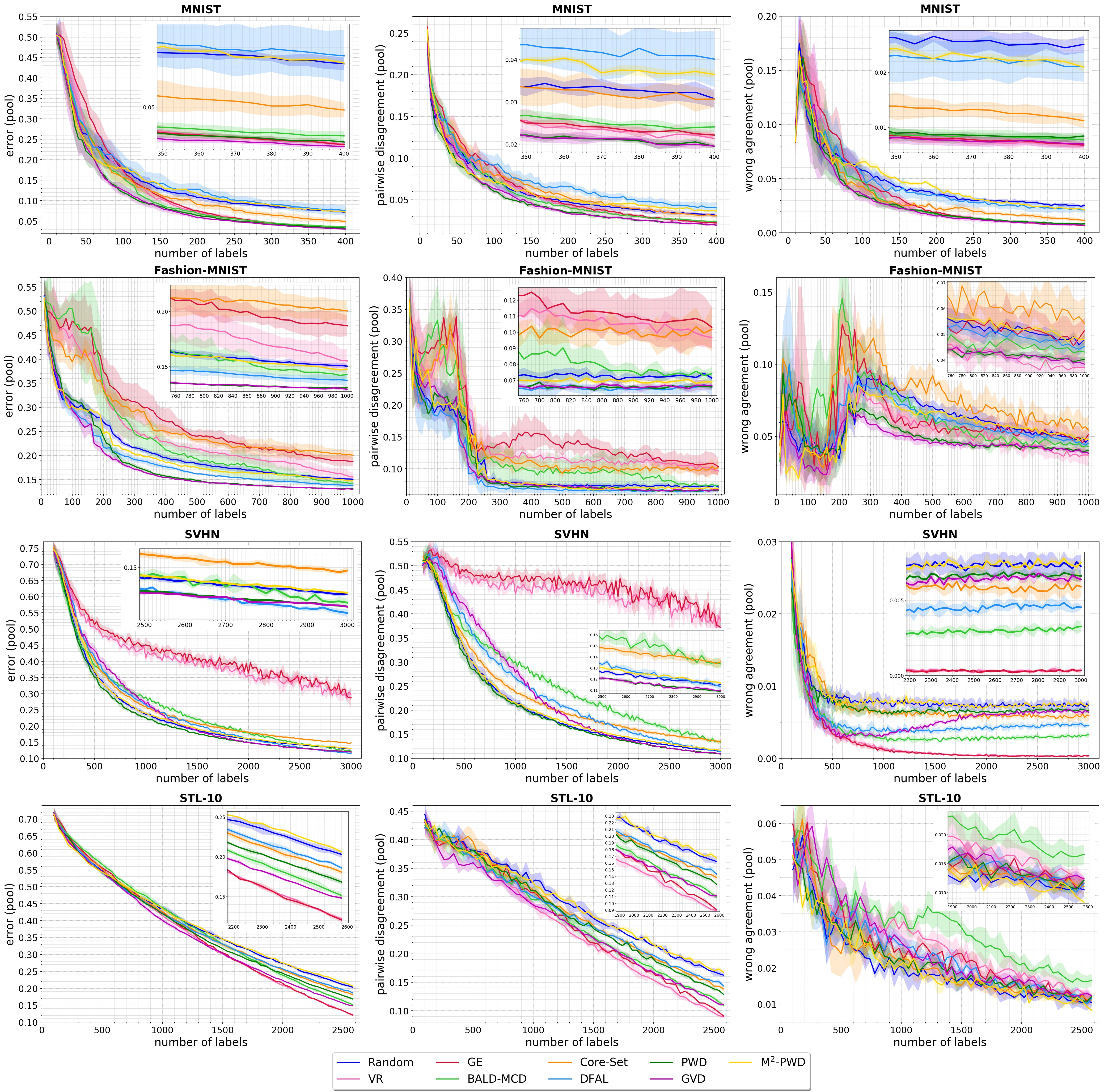}
\end{center}
   \caption{Error rate, pairwise disagreement and wrong agreement over number of queried labels on the pool set.}
\label{fig:err_pw_dis_wrg_agr_pool}
\end{figure*}

In addition to the evaluation results on the test set shown in the paper, we show the results on the pool set as well in Figure~\ref{fig:err_pw_dis_wrg_agr_pool}. On MNIST, Fashion-MNIST and SVHN, diameter reduction are more effective than prior mass reduction methods and other baselines in finding the true labeling on the pool set using as few labels as possible. On STL-10, prior mass reduction performs better at reducing the diameter. An explanation is that we use unlabeled samples in the validation set to estimate relative prior mass and diameters when selecting each query. So the diameter reduction methods are not explicitly optimizing the diameter measured on the pool set, but rather on an unbiased validation set.  

\subsection{Embedding of Version Spaces}
\begin{figure*}[ht!]
\begin{center}
\includegraphics[width=1.0\linewidth]{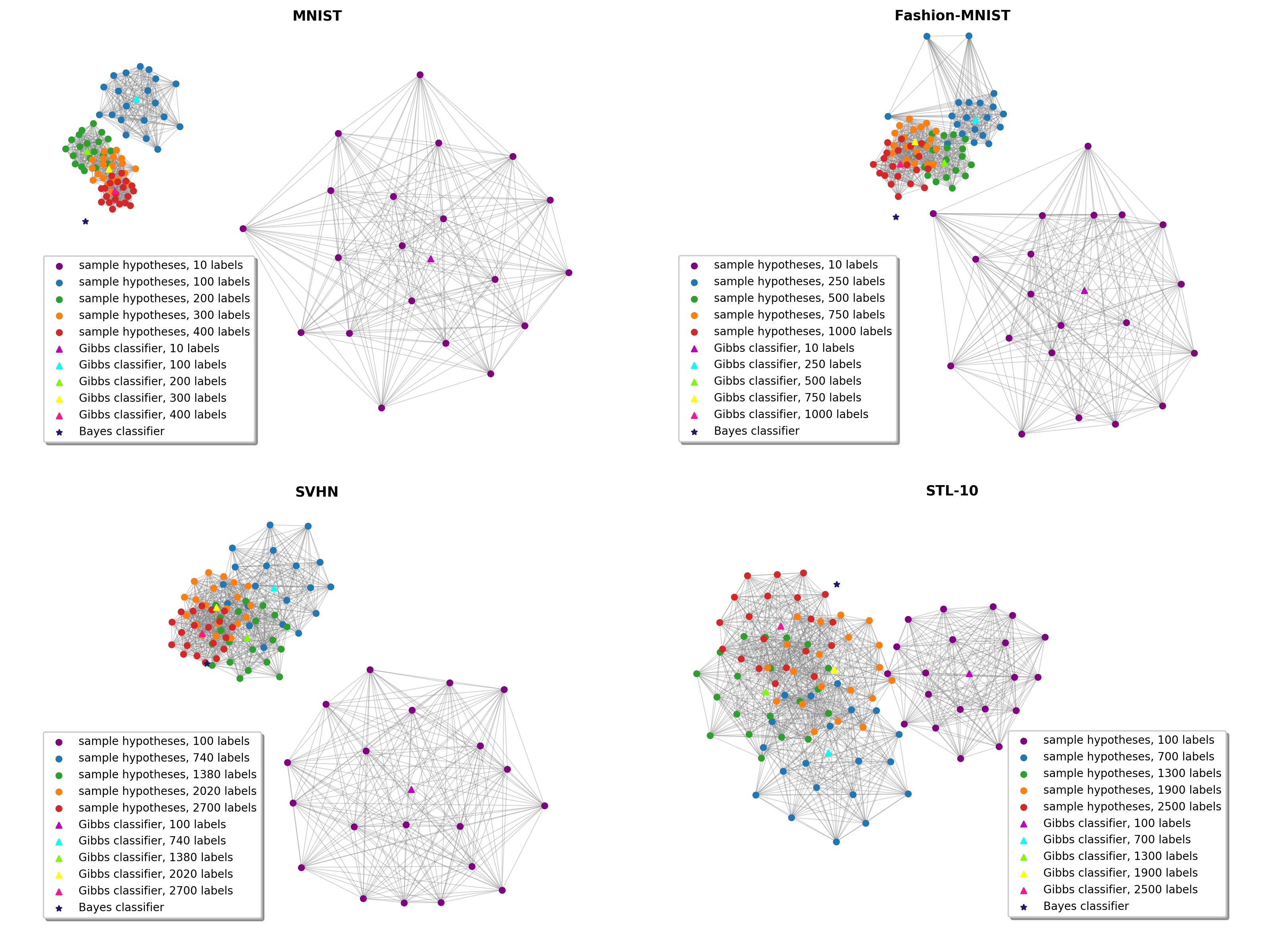}
\end{center}
   \caption{Embedding of version spaces using Multi-Dimensional Scaling (MDS). The 2-D embedding of version spaces obtained by training on random samples is calculated through MDS for each dataset. The purple dots in the largest clusters illustrate a set of sample hypotheses from the version space at the beginning of the active learning experiments, while the red dots in the smallest clusters illustrate hypotheses at the end, being closer to the Bayes classifier (star marker) than those from other version spaces. The blue, green and orange dots represent version spaces obtained by training with (approximately) 25\%, 50\% and 75\% labels of the total budget, respectively. The Gibbs classifier (triangle marker) corresponding to each version space is a random classifier that predicts by randomly sampling a hypothesis from the version space and making the same prediction as the sample hypothesis does.}
\label{fig:vspace_mds}
\end{figure*}

To better illustrate the evolution of version spaces and the existence of bias in version spaces, we show a 2-D embedding of sample hypotheses during the active learning process for each dataset using Multi-Dimensional Scaling (MDS) \cite{Kruskal78} in Figure~\ref{fig:vspace_mds}. MDS finds a low-dimensional representation of potentially high-dimensional data by preserving pairwise distances between the data points. We show sample hypotheses from the first (purple) and the last (red) version spaces as well as intermediate version spaces obtained by training on randomly queried labels (approximately) amount to 25\%, 50\% and 75\% of the total budget. To achieve better visualization, we first compute the embedding of the five Gibbs (random) classifiers and the Bayes classifier and then compute the embedding of each version space separately and center them at the corresponding Gibbs classifier. We use the disagreement probability evaluated on the test set as the distance metric.

As more labels are queried, the version spaces move closer to the Bayes classifier while reducing their diameters. The bias in the version spaces is non-negligible for the four datasets. An active learning algorithm contributes to the shrinkage of the samplable version space but does not have direct control over the reduction of bias. How to efficiently reduce the bias remains an open problem for designing active learning algorithms for neural networks.

\section{Datasets Selection}
The four image classification datasets MNIST, Fashion-MNIST, SVHN and STL-10 are chosen based on several considerations: (1) relatively balanced label distribution; (2) there exist neural network models that can train fast on the original training set of the datasets; (3) no data augmentations are needed. Since active learning methods query highly biased samples, a balanced label distribution help mitigate the problem of query label imbalance. The second point helps reduce the time needed to run active learning experiments. The last point guarantees that the samples used for training are exactly those that have been queried.

\section{Implementation Details}
For the four datasets we consider, no data augmentation is used. Unless otherwise stated, the neural network models are trained using SGD with initial learning rate $1\mathrm{e}{-2}$ and momentum 0.8. The learning rate decays by a factor of 0.1 when there are no improvements on the validation accuracy for any consecutive 10 training epochs until it is smaller than $1\mathrm{e}{-4}$. The maximum training epochs are 200. The batch size for training is set to 32. 

\subsection{MNIST \cite{Lecun10} }
We select a random balanced set of 50000 samples from the original 60000 training samples as the training/validation set and use the original 10000 test samples as the test set. The 2-conv-layer ConvNet is trained using RMSProp \cite{Tieleman12} with learning rate $1\mathrm{e}{-4}$. The learning rate decays by a factor of 0.5 when there are no improvements on the validation accuracy for any consecutive 10 training epochs until it is smaller than $1\mathrm{e}{-5}$. Dropout \cite{Srivastava14} of rate 0.5 is applied to the output of the fully-connected layer which lies between the last convolution layer and the output layer. The batch size for training is set to 16.

\subsection{Fashion-MNIST \cite{Xiao17} }
We used the original balanced 60000 training and 10000 test samples as the training/validation and test sets. A 3-layer-conv ConvNet is used as the classifier model. Dropout of rate 0.5 is applied to the output of the fully-connected layer which lies between the last convolution layer and the output layer. 

\subsection{SVHN \cite{Netzer11} }
We select a random balanced set of 45000 samples from the original 73257 training samples as the training/validation set and a random balanced 15000 samples from the original 26032 test samples as the test set. A 6-layer-conv ConvNet is used as the classifier model. Dropouts of rate 0.3 are applied to the output of every two convolution layers and that of the fully-connected layer which lies between the last convolution layer and the output layer. 

\subsection{STL-10 \cite{Coates11} }
We used the original balanced 5000 training and 8000 test samples as the training/validation and test sets. ResNet18 \cite{He16} is used as the classifier model. Dropout of rate 0.5 is applied between all convolution layers in each convolutional block \cite{Zagoruyko16}.

\bibliographystyle{splncs04}
\bibliography{./references}

\end{document}